\documentclass[letterpaper]{article} 
\usepackage{aaai25}  
\usepackage{times}  
\usepackage{helvet}  
\usepackage{courier}  
\usepackage[hyphens]{url}  
\usepackage{graphicx} 
\usepackage{natbib}  
\usepackage{caption} 
\usepackage{algorithm}
\usepackage{algorithmic}
\usepackage{xcolor}

\usepackage{newfloat}
\usepackage{listings}
\DeclareCaptionStyle{ruled}{labelfont=normalfont,labelsep=colon,strut=off} 
\floatstyle{ruled}
\newfloat{listing}{tb}{lst}{}
\floatname{listing}{Listing}
\title{MathMistake Checker: A Comprehensive Demonstration for Step-by-Step Math Problem Mistake Finding by Prompt-Guided LLMs}
\author {
    Tianyang Zhang\textsuperscript{\rm 1}$^*$,
    Zhuoxuan Jiang\textsuperscript{\rm 2}\thanks{The authors contributed equally to this work. Zhuoxuan Jiang is the corresponding author.},
    Haotian Zhang\textsuperscript{\rm 1}$^*$,
    Lin Lin\textsuperscript{\rm 3},
    Shaohua Zhang\textsuperscript{\rm 2}
}
\affiliations {
    \textsuperscript{\rm 1}Learnable.ai, Shanghai, China\\
    \textsuperscript{\rm 2}Shanghai Business School, Shanghai, China\\
    \textsuperscript{\rm 3}UCloud Technology Co. Ltd., Shanghai, China\\
    \{tianyang.zhang,haotian.zhang\}@learnable.ai, \{jzx,zhangsh\}@sbs.edu.cn, lesley.lin@ucloud.cn
}

\begin{document}

\maketitle

\begin{abstract}
We propose a novel system, MathMistake Checker, designed to automate step-by-step mistake finding in mathematical problems with lengthy answers through a two-stage process. The system aims to simplify grading, increase efficiency, and enhance learning experiences from a pedagogical perspective. It integrates advanced technologies, including computer vision and the chain-of-thought capabilities of the latest large language models (LLMs). Our system supports open-ended grading without reference answers and promotes personalized learning by providing targeted feedback. We demonstrate its effectiveness across various types of math problems, such as calculation and word problems.
\end{abstract}

\section{Introduction}

In mathematics education, accurately assessing students' problem-solving methods has long been a critical challenge, particularly for word problems that require step-by-step verification of a student's answer~\cite{black1998assessment, yavuz2024utilizing}. Unlike multiple-choice or fill-in-the-blank questions that demand only brief responses, effectively assessing the reasoning between adjacent steps in longer answers is pedagogically significant, as it measures student understanding, informs instruction, and provides meaningful feedback~\cite{hattie2007power}.
 
Checking each step of student answers is a challenging task. Traditionally, it has been time-consuming for human teachers to read every answer step when grading student homework or examinations. Reference answers are often helpful for teachers to quickly identify mistakes and expedite the grading process. Consequently, some automated grading systems have been developed to verify student answers against a reference answer.

However, traditional grading systems have inherent limitations. They typically depend on predefined reference answers, which restrict their ability to recognize and evaluate the diverse approaches students may employ. This can result in a narrow interpretation of student competence, overlooking alternative methods and mathematically valid creative solutions. As educational practices shift toward promoting higher-order thinking and problem-solving skills, there is an increasing need for grading systems that can manage this complexity, providing more detailed and flexible assessments that align with the various ways students approach problems~\cite{black2009developing,popham2009assessment}.

\begin{figure}
\centerline{\includegraphics[width=0.46\textwidth]{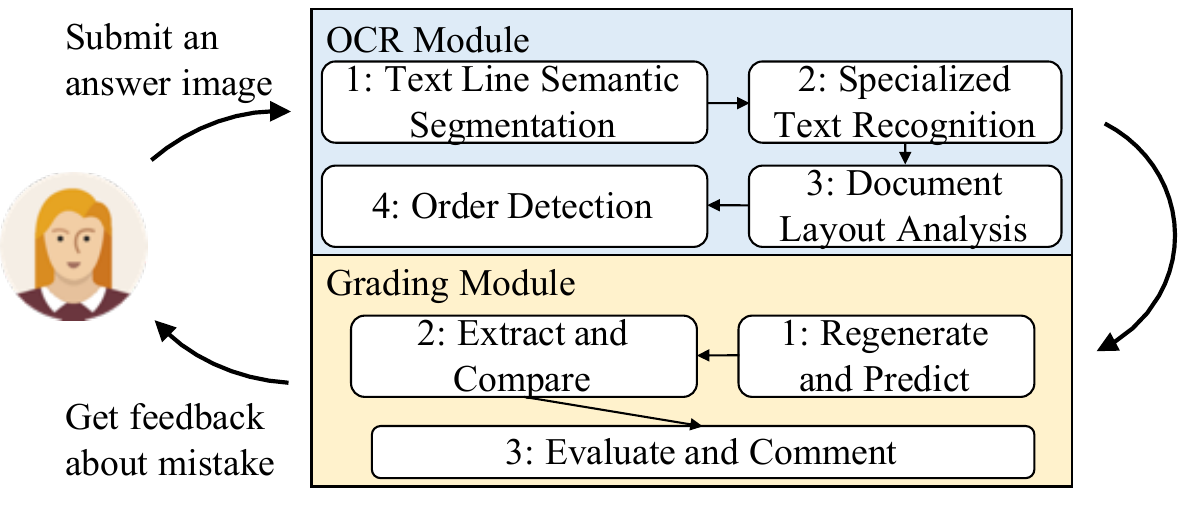}}
\caption{Architecture of MathMistake Checker.} \label{1}
\end{figure}

To develop a grading system that operates without reference answers, two essential capabilities are required: Optical Character Recognition (OCR) and decision-making for mistake finding. The former handles the interpretation of image-based inputs, while the latter focuses on grading the answers. With the recent advancements in large models, Large Vision-Language Models (LVLMs)~\cite{adewumi2024fairness} can be utilized for end-to-end grading of mathematical answers. However, current LVLMs primarily emphasize visual understanding, often overlooking the reasoning aspect of natural language, particularly when mathematical texts demand more precise multi-modal comprehension. In this paper, we propose a two-stage approach to enhance interpretability and facilitate model training. As shown in Figure~\ref{1}, the OCR Module in Stage 1 involves transforming image-based inputs into textual information, while the Grading Module in Stage 2 focuses on step-by-step mistake finding.

Specifically, we propose a demo system called MathMistake Checker, which integrates several cutting-edge technologies. In Stage 1, the system focuses on the precise extraction and correction of mathematical content from images submitted by users. This stage includes a pipeline with several phases: text line semantic segmentation~\cite{chen2018encoder,xie2021segformer}, specialized text recognition~\cite{shi2016end,li2023trocr}, document layout analysis~\cite{huang2022layoutlmv3}, and order detection~\cite{vinyals2015pointer}. In Stage 2, we utilize the latest Large Language Models (LLMs) to serve as the step-by-step grader for the recognized text. Building on existing research that highlights LLMs' reasoning capabilities~\cite{ishida2024large,lundgren2024large,xie2024grade}, we specifically incorporate the Pedagogical Chain-of-Thought (PedCoT) prompting strategy~\cite{jiang2024llms}, which effectively identifies logical mistakes in students' answers step-by-step. Notably, our system can support a variety of LLMs and prompting strategies.

Our demo system provides detailed feedback on the student’s steps by systematically analyzing their problem-solving approach, identifying key steps and potential missteps, and providing targeted feedback that addresses specific areas of misunderstanding or mistake. It is designed to enhance pedagogical significance by aligning with the student's thought process and providing feedback that is both relevant and constructive.

\section{System Overview}
The proposed two-stage MathMistake Checker comprises two primary modules: OCR module and grade module.

\subsection{OCR Module}

The OCR Module employs a multiphase pipeline to process handwritten mathematical contents.

\subsubsection{Text Line Semantic Segmentation}

This phase receives raw images with printed questions and handwritten answers, using semantic segmentation models~\cite{chen2018encoder,xie2021segformer} to separate text lines into printed text, handwritten text, and equations. This segmentation enhances accuracy in subsequent steps.

\subsubsection{Specialized Text Recognition}

During this phase, each segmented text line is processed individually, with the appropriate model selected based on content type. Vision encoder-decoder transformer-based models~\cite{li2023trocr} are used for handwritten text and equations, while a convolutional recurrent neural network~\cite{shi2016end} handles printed text. This multi-model approach ensures high throughput and precision across diverse text formats.

\subsubsection{Document Layout Analysis}

After recognizing text lines, a multi-modal transformer~\cite{huang2022layoutlmv3} is used for document layout analysis to understand spatial and logical relationships. This step reconstructs structures like tables and diagrams, preserving their original format.

\subsubsection{Order Detection}

This phase is for reconstructing the original writing sequence of the handwritten content. By utilizing a Pointer Network~\cite{vinyals2015pointer}, the correct order of text lines based on positional data can be determined, particularly in free-form layouts. This ensures an accurate logical step flow of the student's responses.

\subsection{Grading Module}

The Grading Module functions in three phases to evaluate a student's answer by using LLMs. For details on prompt strategies, refer to the original papers~\cite{jiang2024llms}.

\subsubsection{Regenerate and Predict}

The system prompts the selected LLM with the problem statement and the initial steps of the student's solution using a specified prompting strategy. The LLM then generates the expected mathematical concepts, problem-solving approaches, and calculations for the next step without accessing the student’s actual response.

\subsubsection{Extract and Compare}

In this phase, the LLM receives the student's actual response. It extracts the key elements of the answer and compares them with the predictions from previous phase. The system labels any discrepancies as mistakes, categorizing them according to predefined criteria such as correctness and alignment.

\subsubsection{Evaluate and Comment}

In this phase, the system prompts the LLM to evaluate the student's reasoning up to the current step. The LLM provides a concise assessment, indicating whether the current step is correct, incorrect, or partially correct, and offers targeted feedback on the student's approach.

\begin{figure}
\centerline{\includegraphics[width=0.46\textwidth]{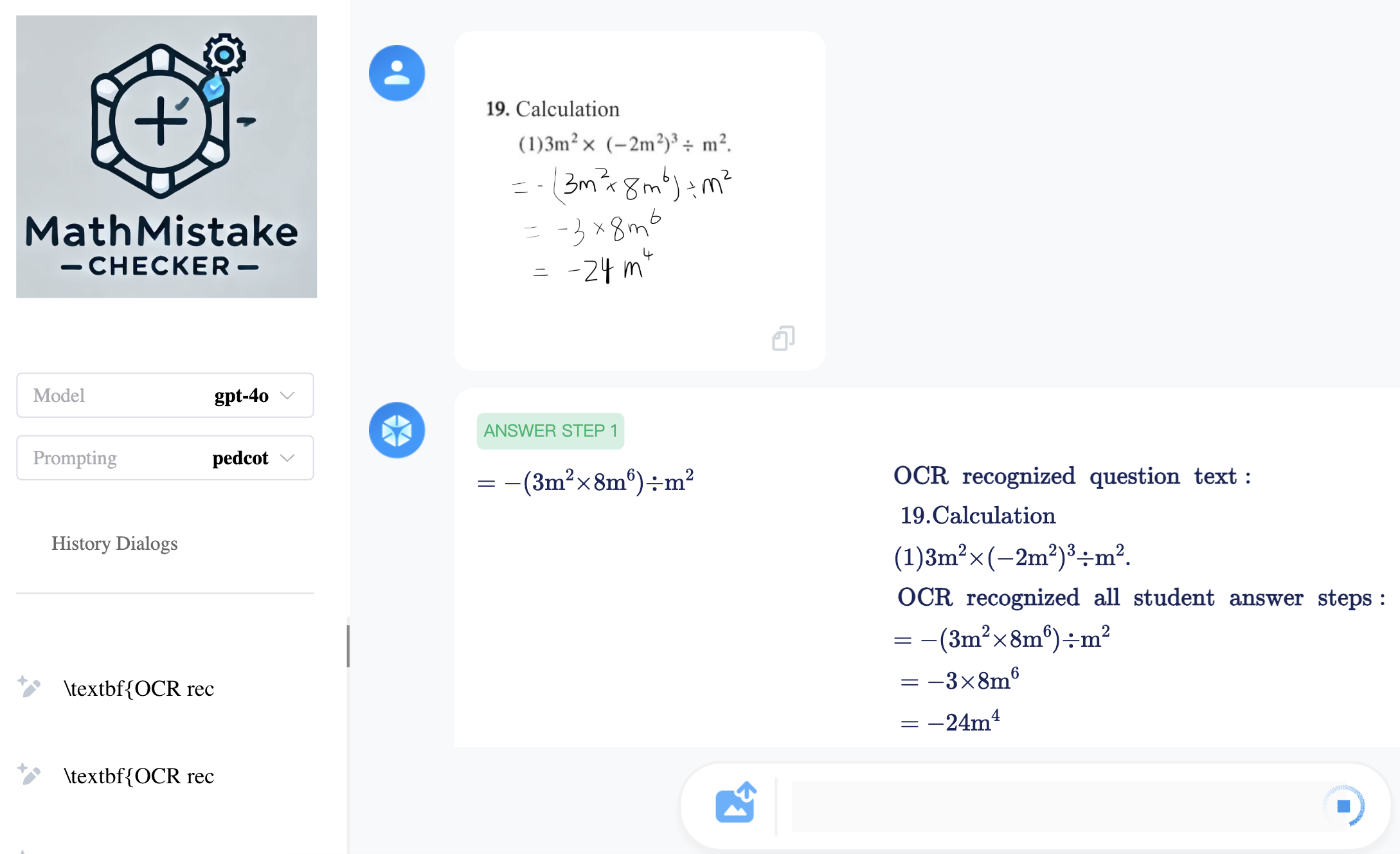}}
\caption{Demo interface for grading the math answer.} \label{2}
\end{figure}

\section{Case Studies}

We apply our demo system to several cases. Figure~\ref{2} illustrates the demo interface for grading the answer to a math calculation problem. The left section of drop-down boxes provides options for various LLMs and prompting strategies, while the right section features a dialog interaction window that displays the step-by-step grading process. For more detailed information about the demos, please refer to the video.

\section{Conclusion}

In this paper, we present MathMistake Checker, a demo system that automates the grading of mathematical problems by integrating advanced technologies such as OCR and the chain-of-thought approach in the latest large language models (LLMs). The system delivers accurate and open-ended grading while supporting personalized learning by providing targeted feedback. MathMistake Checker streamlines the grading process and enhances learning experiences from a pedagogical perspective. Future work will focus on extending the system’s capabilities to other subjects and improving the quality of feedback.

\section{Acknowledgments}
This work is partially supported by International Science and Technology Cooperation Program of Shanghai (No. 24170790602). We thank all the anonymous reviewers for their insightful and constructive comments.

\bibliography{aaai25}

\begin{thebibliography}{16}
\providecommand{\natexlab}[1]{#1}

\bibitem[{Adewumi et~al.(2024)Adewumi, Alkhaled, Gurung, van Boven, and
  Pagliai}]{adewumi2024fairness}
Adewumi, T.; Alkhaled, L.; Gurung, N.; van Boven, G.; and Pagliai, I. 2024.
\newblock Fairness and bias in multimodal ai: A survey.
\newblock \emph{arXiv preprint arXiv:2406.19097}.

\bibitem[{Black and Wiliam(1998)}]{black1998assessment}
Black, P.; and Wiliam, D. 1998.
\newblock Assessment and classroom learning.
\newblock \emph{Assessment in Education: principles, policy \& practice}, 5(1):
  7--74.

\bibitem[{Black and Wiliam(2009)}]{black2009developing}
Black, P.; and Wiliam, D. 2009.
\newblock Developing the theory of formative assessment.
\newblock \emph{Educational Assessment, Evaluation and Accountability
  (formerly: Journal of personnel evaluation in education)}, 21: 5--31.

\bibitem[{Chen et~al.(2018)Chen, Zhu, Papandreou, Schroff, and
  Adam}]{chen2018encoder}
Chen, L.-C.; Zhu, Y.; Papandreou, G.; Schroff, F.; and Adam, H. 2018.
\newblock Encoder-decoder with atrous separable convolution for semantic image
  segmentation.
\newblock In \emph{Proceedings of the European conference on computer vision
  (ECCV)}, 801--818.

\bibitem[{Hattie and Timperley(2007)}]{hattie2007power}
Hattie, J.; and Timperley, H. 2007.
\newblock The power of feedback.
\newblock \emph{Review of educational research}, 77(1): 81--112.

\bibitem[{Huang et~al.(2022)Huang, Lv, Cui, Lu, and Wei}]{huang2022layoutlmv3}
Huang, Y.; Lv, T.; Cui, L.; Lu, Y.; and Wei, F. 2022.
\newblock Layoutlmv3: Pre-training for document ai with unified text and image
  masking.
\newblock In \emph{Proceedings of the 30th ACM International Conference on
  Multimedia}, 4083--4091.

\bibitem[{Ishida et~al.(2024)Ishida, Liu, Wang, and Cheung}]{ishida2024large}
Ishida, T.; Liu, T.; Wang, H.; and Cheung, W.~K. 2024.
\newblock Large Language Models as Partners in Student Essay Evaluation.
\newblock \emph{arXiv preprint arXiv:2405.18632}.

\bibitem[{Jiang et~al.(2024)Jiang, Peng, Feng, Li, and Li}]{jiang2024llms}
Jiang, Z.; Peng, H.; Feng, S.; Li, F.; and Li, D. 2024.
\newblock LLMs can Find Mathematical Reasoning Mistakes by Pedagogical
  Chain-of-Thought.
\newblock \emph{arXiv preprint arXiv:2405.06705}.

\bibitem[{Li et~al.(2023)Li, Lv, Chen, Cui, Lu, Florencio, Zhang, Li, and
  Wei}]{li2023trocr}
Li, M.; Lv, T.; Chen, J.; Cui, L.; Lu, Y.; Florencio, D.; Zhang, C.; Li, Z.;
  and Wei, F. 2023.
\newblock Trocr: Transformer-based optical character recognition with
  pre-trained models.
\newblock In \emph{Proceedings of the AAAI Conference on Artificial
  Intelligence}, 13094--13102.

\bibitem[{Lundgren(2024)}]{lundgren2024large}
Lundgren, M. 2024.
\newblock Large Language Models in Student Assessment: Comparing ChatGPT and
  Human Graders.
\newblock \emph{arXiv preprint arXiv:2406.16510}.

\bibitem[{Popham(2009)}]{popham2009assessment}
Popham, W.~J. 2009.
\newblock Assessment literacy for teachers: Faddish or fundamental?
\newblock \emph{Theory into practice}, 48(1): 4--11.

\bibitem[{Shi, Bai, and Yao(2016)}]{shi2016end}
Shi, B.; Bai, X.; and Yao, C. 2016.
\newblock An end-to-end trainable neural network for image-based sequence
  recognition and its application to scene text recognition.
\newblock \emph{IEEE transactions on pattern analysis and machine
  intelligence}, 39(11): 2298--2304.

\bibitem[{Vinyals, Fortunato, and Jaitly(2015)}]{vinyals2015pointer}
Vinyals, O.; Fortunato, M.; and Jaitly, N. 2015.
\newblock Pointer networks.
\newblock \emph{Advances in neural information processing systems}, 28.

\bibitem[{Xie et~al.(2021)Xie, Wang, Yu, Anandkumar, Alvarez, and
  Luo}]{xie2021segformer}
Xie, E.; Wang, W.; Yu, Z.; Anandkumar, A.; Alvarez, J.~M.; and Luo, P. 2021.
\newblock SegFormer: Simple and efficient design for semantic segmentation with
  transformers.
\newblock \emph{Advances in neural information processing systems}, 34:
  12077--12090.

\bibitem[{Xie et~al.(2024)Xie, Niu, Xue, and Guan}]{xie2024grade}
Xie, W.; Niu, J.; Xue, C.~J.; and Guan, N. 2024.
\newblock Grade Like a Human: Rethinking Automated Assessment with Large
  Language Models.
\newblock \emph{arXiv preprint arXiv:2405.19694}.

\bibitem[{Yavuz, {\c{C}}elik, and
  Yava{\c{s}}~{\c{C}}elik(2024)}]{yavuz2024utilizing}
Yavuz, F.; {\c{C}}elik, {\"O}.; and Yava{\c{s}}~{\c{C}}elik, G. 2024.
\newblock Utilizing large language models for EFL essay grading: An examination
  of reliability and validity in rubric-based assessments.
\newblock \emph{British Journal of Educational Technology}.

\end{thebibliography}

\end{document}